%% file: main.tex
\documentclass[runningheads]{llncs}

\usepackage[year=2026,ID=9]{eccv}
\usepackage{eccvabbrv}

\usepackage{lmodern}
\usepackage[T1]{fontenc}
\usepackage{amsmath,amssymb}
\usepackage{booktabs}
\usepackage{graphicx}
\usepackage{hyperref}
\usepackage{xcolor}
\usepackage{xspace}
\usepackage{cite}
\usepackage{hyperref}
\usepackage{pifont}
\newcommand{\chg}[1]{\,(#1)}

\renewcommand{\baselinestretch}{0.95}

\begin{document}

\title{Signpost Watermarking: Joint Optimization for Visual Watermark Coexistence}
%\title{Signpost Watermarking: Optimizing Visual Watermark Coexistence for Content Provenance}

\titlerunning{Signpost Watermarking}

% Include the authors' ORCID for the camera-ready version, if at all possible.
\author{Shruti Agarwal\inst{1} \and Vishal Asnani\inst{1} \and John Collomosse\inst{1,2}}

\authorrunning{S.~Agarwal, V.~Asnani and J.~Collomosse}

\institute{Adobe Research, San Jose CA 95110, USA \and University of Surrey, Guildford GU2 7XH, UK}

\maketitle

\begin{abstract}

We present a method for training imperceptible visual watermarks to coexist with other such watermarks. Recent work has shown that independently trained image watermarking models can coexist with surprisingly limited interference, enabling watermark ensembling. However, this coexistence is a serendipitous property rather than an explicit optimization objective, leaving interference uncontrolled and potentially reducing decoding robustness or visual quality. We first show empirically that the same coexistence property extends to video watermarking. We then show that both image and video watermarks can be trained with a decoder-aware objective to improve coexistence. Our results suggest a practical path to signpost watermarks that indicate the presence of independently deployed provenance watermarking systems, supporting layered provenance signaling for content authenticity and rights.

\end{abstract}

\renewcommand{\baselinestretch}{0.98}
\input{sections/01_intro.tex}
\input{sections/02_related.tex}
\input{sections/03_method.tex}

\input{sections/04_experiments.tex}
\input{sections/05_conclusion.tex}

\section{Acknowledgement}
 This work was supported in part by UKRI/EPSRC grant DECaDE EP/T022485/1.  We appreciate many discussions on signpost watermarking with Andy Parsons.
 
\bibliographystyle{splncs04}
\bibliography{main.bib}

\end{document}

%% file: sections/01_intro.tex
\section{Introduction}
\label{sec:intro}

Imperceptible digital watermarking is increasingly being adopted to address questions of content rights and transparency, particularly in the context of generative AI. Regulators (\eg via the EU AI Act, Article 50 \cite{eu_ai_act_2024}), standards bodies (\eg C2PA \cite{c2pa}), and technology providers (\eg Google's SynthID \cite{synthid}, Meta's VideoSeal \cite{fernandez2024video}, Adobe's TrustMark \cite{bui2025trustmark}) have all advocated for watermarking systems that can disclose the use of generative AI or support provenance tracking for digital media.

At a technical level, watermarking embeds information, or a `payload', into media content such that it can later be recovered while remaining largely imperceptible to human observers. This payload may serve as a binary label indicating that AI was used in the creation of content, or as a unique asset identifier that references provenance metadata describing origin, authorship, or intellectual property (IP) rights. The diversity of watermarking use cases, coupled with the practical need to keep many watermarking systems secret to reduce attacks, has led to a proliferation of independently developed and non-interoperable watermarking technologies, particularly in the visual domain of images and video.  Yet there is an urgent need to encourage interoperability in visual watermarking systems given the importance of content authenticity and IP rights.

Recent work by Petrov et al.~\cite{Petrov-NeurIPS-2025} demonstrated that independently developed watermarking systems can coexist within the same {\em image} with surprisingly limited interference, enabling watermark ensembling and improved robustness--capacity--quality trade-offs. However, existing approaches treat watermark coexistence as an emergent property of independently trained watermarking systems rather than as an optimization objective. As a result, interference between coexisting watermarks remains uncontrolled and may reduce decoding robustness or visual quality. The question follows: {\em can visual watermarking algorithms be explicitly trained to improve their coexistence properties within the same asset?}

\textbf{Signpost Watermarking.} The ability for multiple watermarks to coexist within a single image or video creates opportunities for interoperability through a lightweight routing, or `signpost', watermark. Such a signpost serves only to indicate the presence of a coexisting watermark that carries an application-level signal, such as an AI label or provenance identifier \eg that carries authenticity or IP rights information. The signpost therefore supports efficient discovery of the proprietary watermarking system used to carry that signal, avoiding the $1$-to-$N$ problem of attempting to decode against all $N$ watermarking technologies commonly used in a given modality. This requires only standardizing how to encode and detect the signpost watermark, rather than standardizing a single watermarking system for all use cases, which is unlikely to be practical for technical or commercial reasons. An alternative would be to train a shared classifier that predicts which of $N$ watermarking technologies is present within a media asset. However, such a classifier would, by design, require representative watermarked training data and not naturally extend to the introduction of an $(N+1)^\mathrm{th}$ technology. By contrast, a signpost watermark can be extended through a simple registry or list that enumerates the watermarking methods expressible in the signpost payload.

This paper makes two technical contributions in support of signpost watermarking for visual content.

\noindent\textbf{1. Video watermark coexistence.}
We demonstrate that the emergent coexistence properties observed for common image watermarks also hold for common video watermarks. To our knowledge, no prior work has investigated coexistence among independently developed video watermarking systems.

\noindent\textbf{2. Decoder-aware training.}
We show that coexistence can be treated as a training objective. If a watermark is intended to coexist with other watermarking systems, it should be optimized explicitly to preserve their decoding performance. This perspective transforms coexistence from an empirical observation into a multi-objective learning problem. Given a collection of frozen watermark decoders, we show that a signpost encoder--decoder can be optimized not only for signpost robustness and perceptual quality, but also to preserve the decoding accuracy of coexisting watermarking systems.

We evaluate our decoder-aware signpost training approach across up to four state-of-the-art image watermarking systems and three video watermarking systems. We analyze how coexistence changes as the co-present watermark technologies are added or removed from the training, and evaluate robustness under a diverse set of transformations. Our results show decoder-aware optimization significantly improves coexistence while maintaining strong signpost robustness and perceptual quality.

%% file: sections/02_related.tex
\section{Related Work}
\label{sec:related}

\paragraph{Image watermarking.}
Digital watermarking has long been studied as a mechanism for embedding imperceptible signals into images. Classical approaches typically embedded signals in spatial or frequency domains, while recent neural methods learn encoder--decoder pairs that optimize payload recovery, imperceptibility, and robustness jointly. Methods such as HiDDeN~\cite{zhu2018hidden}, StegaStamp~\cite{tancik2020stegastamp}, RoSteALS~\cite{bui2023rosteals}, TrustMark~\cite{bui2025trustmark}, and related systems train watermarks to survive common image transformations including compression, resizing, cropping, noise, and color changes. Parallel work has considered watermarking generative model outputs directly, including diffusion-model watermarks such as Tree-Ring~\cite{wen2023treerings}, Stable Signature~\cite{fernandez2023stable}, and proprietary commercial systems such as SynthID~\cite{synthid}. Recent work by Petrov et al.~\cite{Petrov-NeurIPS-2025} showed that independently trained image watermarking systems can coexist. Our work builds on this observation by asking whether coexistence can be extended to video and optimized directly during training.

\paragraph{Video watermarking.}
Video watermarking extends image watermarking to the temporal domain, where embedded signals must remain imperceptible across frames while surviving spatial transformations, temporal processing, and video compression. Classical methods can be broadly grouped into codec-integrated and spatial-domain approaches. Codec-integrated techniques embed signals directly into compressed streams, for example by modifying DCT coefficients, entropy codes, or motion vectors in MPEG-2, H.264/AVC, or H.265/HEVC~\cite{Yong2008,Alattar2003,Zhang2007,Guo2010,Tew2016,Dutta2018}. These methods are efficient and can offer high visual quality, but are often tightly coupled to specific codecs and fragile under transcoding. Recent deep-learning approaches such as VStegNet~\cite{mishra2019vstegnet} and RivaGAN~\cite{RivaGAN2019} extended neural image watermarking to video, with RivaGAN introducing an attention-based architecture for robust invisible video watermarking. Differentiable compression modules and codec-aware training have been explored to improve robustness and perceptual quality~\cite{zhang2023novel,shen2023vhnet,zhang2024v2a}. VideoSeal~\cite{fernandez2024video} provides an efficient per-frame neural video watermarking model, blending watermark residuals to reduce flicker. Recently, FlowMark~\cite{asnani2026flowmark} propose a mask-based video watermarking technique with a temporal loss to reduce flicker and improve watermark quality. However, existing video watermarking systems are typically trained and evaluated in isolation; whether independently developed video watermarks can coexist within the same video, and whether such coexistence can be improved through training, are both unexplored issues.

\paragraph{Media provenance and watermark interoperability.}

Media provenance systems aim to record the origin, authorship, editing history, and rights associated with digital content. Early works such as ARCHANGEL\cite{Collomosse-DocEng-2018,Bui-CVPRWS-2019}, and later AMP\cite{AMP-2021}, stored  cryptographically signed provenance data as tamper-evident records on a Blockchain. Today, open standards such as C2PA~\cite{c2pa} store provenance data through cryptographically signed asset metadata, enabling use cases including media integrity, AI training consent \cite{decorait}, and IP rights~\cite{contentarcs,Collomosse-IEEECGA-2026}. However, asset metadata may be lost through common distribution channels such as screenshots, re-encoding, and social media sharing. Persistent signals such as perceptual hashes and imperceptible watermarks can therefore complement metadata by helping to re-link media assets to provenance records after distribution~\cite{Collomosse2024}. C2PA does not standardize a single watermarking model or payload structure for this purpose, but instead maintains an enumerated list of watermarking techniques in common use. In concurrent work, ZOETROPE \cite{Collomosse-IEEECGA-2026} proposed using a coexisting signpost watermark for images but does not explicitly train it to optimize for coexistence, as we do here, and does not address the video modality.  In this work, we propose training a coexisting signpost watermark to explicitly encode the presence of such watermarking technologies, thereby supporting signpost-based watermark interoperability addressing for images and videos.

%% file: sections/03_method.tex
%Coexistent Signpost: Architecture \& Coexistence Evaluation \\
%{\normalsize Three run-3 variants vs.\ Old Signpost (\texttt{trustmark\_Q\_piat})}\\[3pt]
%{\small 200 images, mirflickr val/19, seed~42, Kornia medium-severity augmentations}

\section{Signpost Watermarking}
\label{sec:architecture}

\begin{figure}[t]
    \includegraphics[width=\textwidth]{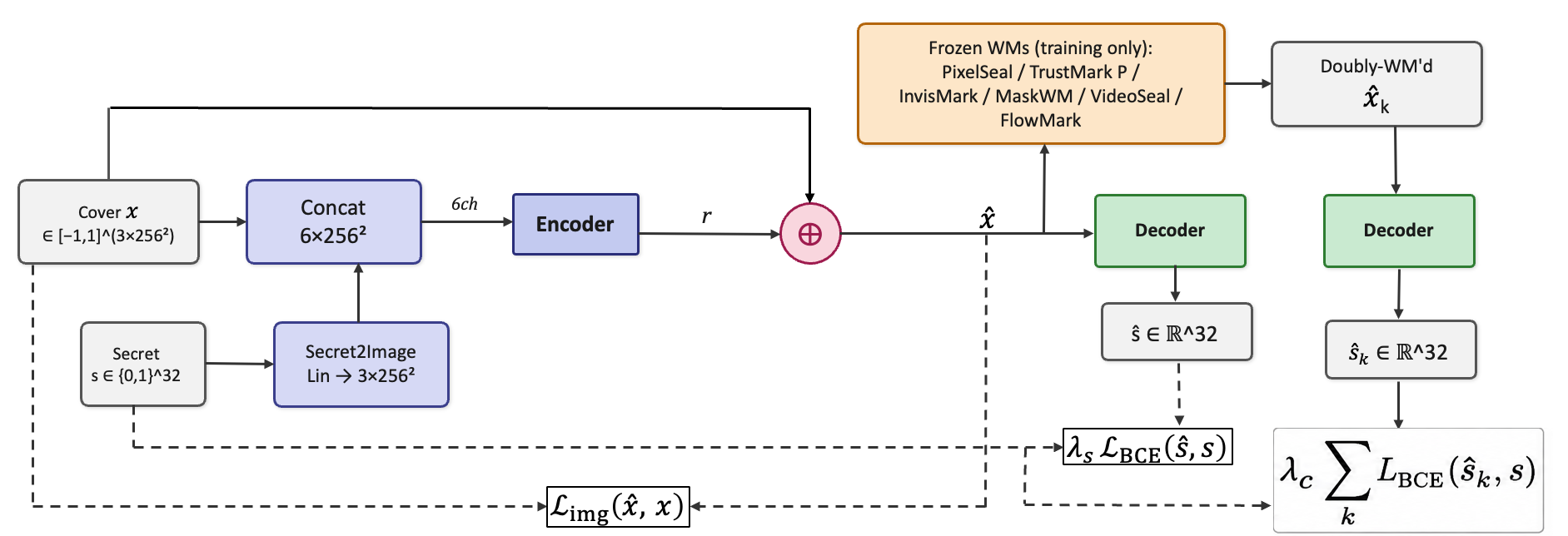}
    \caption{%
  \textbf{Our signpost watermark architecture.}
  \textcolor{blue!60!black}{\textbf{Blue}} = UNet encoder; outputs residual $r$ added to cover $x$ at $\oplus$.
  \textcolor{green!50!black}{\textbf{Green}} = ResNet-50 decoder (recovers 32-bit secret $\hat{s}$ from $\hat{x}$).
  \textcolor{orange!70!black}{\textbf{Orange}} = frozen secondary watermarks applied on top of $\hat{x}$ during training only.
}
\label{fig:arch}
\end{figure}

Figure~\ref{fig:arch} illustrates our signpost architecture, which we use for both image and video experiments.

\paragraph{Encoder.}
The signpost encoder follows a UNet design with 4 downsampling stages and base channel width 32. The 32-bit binary secret $s \in \{0,1\}^{32}$ is first lifted to a spatial representation by a linear projection $\mathbb{R}^{32} \!\to\! \mathbb{R}^{3 \times 16 \times 16}$ followed by nearest-neighbour upsampling to the cover resolution (256$\times$256). This secret map is concatenated channel-wise with the cover image $x$ to form a 6-channel input. Strided convolutions with SiLU activations halve the spatial resolution at each encoding stage; one residual block operates at the bottleneck. Transposed convolutions with skip connections from the corresponding encoder stage reconstruct the spatial resolution, and a final $1\times1$ convolution with tanh activation produces the residual $r \in [-1,1]^{3\times H\times W}$. The stego image is $\hat{x} = \mathrm{clip}(x + r,\,-1,\,1)$. The \emph{nomask} variant does not apply a learnable spatial gating mask to the residual, in contrast to earlier runs that selectively suppressed embedding in perceptually sensitive regions.

\paragraph{Decoder.}
A ResNet-50 backbone maps the stego image (center-cropped or resized to 224$\times$224) to 32 logits. At inference, bit predictions are obtained by sigmoid thresholding at 0.5.

\paragraph{JND guidance.}
A Just Noticeable Difference (JND) map $J\in[0,1]^{H\times W}$ is computed from the luminance of the cover image following a classical spatial-masking model~\cite{wu2017enhanced}. The pixel reconstruction loss is modulated by $(1 - J)$: regions with high JND values (perceptually insensitive, e.g.\ high-frequency textures) incur reduced penalty, allowing the encoder to concentrate embedding energy where distortions are least visible.

\paragraph{Frozen secondary encoders.}
During training, several neural watermarking encoders (cf.~\ref{sec:coexist})  are loaded with frozen weights and applied after (\emph{on top of}) the signpost stego image to provide coexistence supervision. The signpost is applied first as the decode accuracy of the first watermark gets impacted after the second watermark application while the second watermark's decode accuracy remain unaltered~\cite{Petrov-NeurIPS-2025}. 

\paragraph{Training objective.}
The total loss $\mathcal{L}$ combines three terms.  Let $\hat{x}_k$ denote the image obtained by applying frozen watermarker $k$ on top of the signpost stego $\hat{x}$, and let $f_\theta$ denote the signpost decoder.
\begin{equation}
  \mathcal{L} \;=\; \mathcal{L}_{\mathrm{img}}(\hat{x},\,x)
              \;+\; \lambda_s\,\mathcal{L}_{\mathrm{BCE}}\!\bigl(f_\theta(\tilde{x}),\,s\bigr)
              \;+\; \lambda_c\!\sum_k \mathcal{L}_{\mathrm{BCE}}\!\bigl(f_\theta(\hat{x}_k),\,s\bigr)
\end{equation}
where $\tilde{x}$ is $\hat{x}$ after random augmentation noise.  The \emph{image quality} term is
\begin{multline}
  \mathcal{L}_{\mathrm{img}} \;=\; \lambda_r\!\left\|x - \hat{x}\right\|_1
       \;+\; \lambda_r^{\mathrm{YUV}}\!\left\|x - \hat{x}\right\|^2_{\mathrm{YUV}}
       \;+\; \lambda_p\,\mathrm{LPIPS}(x,\,\hat{x}) \\
       \;+\; \lambda_J\,\bigl(\left|x - \hat{x}\right| \odot (1 - J)\bigr),
\end{multline}
combining L1 and YUV-weighted L2 reconstruction with LPIPS perceptual loss~\cite{zhang2018unreasonable} and JND-weighted pixel error.  The last loss, \emph{coexistence} term, is evaluated after each frozen watermark application. This term encourages the signpost encoder to concentrate embedding energy in image components that remain invariant to subsequent watermark overlays. The resulting residual therefore becomes complementary, in both spatial and frequency content, to those of the frozen watermarking systems. 

The signpost survival term $\mathcal{L}_{\mathrm{BCE}}(f_\theta(\tilde{x}), s)$ provides an additional gradient to the decoder, improving its robustness to random augmentations.  Loss weights are $(\lambda_s,\lambda_c)=(20,\,2)$, with $(\lambda_r, \lambda_p, \lambda_J) = (1.5,\,10,\,1)$.

\paragraph{Training curriculum.}
Training follows a three-phase curriculum inspired by TrustMark~\cite{bui2025trustmark}. In \textit{Phase 1}, a fixed cover image is used until signpost bit accuracy exceeds 0.90, establishing a stable embedding signal before the full dataset is introduced. In \textit{Phase 2}, training switches to the full dataset; once bit accuracy exceeds 0.95, augmentation noise is activated (random rotations, crops, Gaussian noise, color jitter, and JPEG compression at increasing severity). In \textit{Phase 3}, once bit accuracy exceeds 0.98 with noise active, image quality losses ramp in over 100,000 steps. The encoder and decoder are jointly optimized with AdamW ($\mathrm{lr}=4{\times}10^{-6}$, weight decay $10^{-4}$, batch size 32).

%% file: sections/04_experiments.tex
\section{Experiments and Discussion}
\label{sec:experiments}

\subsection{Experimental Setup}
\label{sec:setup}

\paragraph{Dataset and evaluation.}
For images, we train on a large-scale proprietary dataset and evaluate on $1000$ images held out from a MirFlickr-1M~\cite{mirflickr,mirflickr1m} validation split. For videos, we trained/tested on 51k/155 from train/val split of SA-V dataset~\cite{ravi2024sam}. Each version is trained at $256\times256$ resolution.

\paragraph{Baselines.}

For the \emph{image} signpost we train against four image watermarking systems:
PixelSeal (PS)~\cite{souvcek2025pixel}, InvisMark (IM)~\cite{xu2025invismark},
TrustMark-P (TM-P)~\cite{bui2025trustmark}, and MaskWM (MW)~\cite{hu2026mask}. We additionally
compare against a prior signpost baseline (ZOETROPE~\cite{Collomosse-IEEECGA-2026}) trained for images, without
the coexistence objective or JND guidance. Because it must survive video-specific
distortions, the \emph{video} signpost is trained separately: we replace PS, TM-P,
and MW with VideoSeal (VS)~\cite{fernandez2024video}, TrustMark-Q (TM-Q)~\cite{bui2025trustmark}, and
FlowMark (FM)~\cite{asnani2026flowmark}, retaining InvisMark as the image watermark, and also
compare against a signpost built on the FlowMark architecture \emph{without} the coexistence
objective (FM-32).

\paragraph{Metrics.}
\textbf{PSNR} (dB, $\uparrow$) measures perceptual quality of the watermarked image against the clean original. For videos, we additionally report VMAF\cite{netflix2019video}, a perceptual quality metric that correlates closely with human judgment.
\textbf{Bit accuracy} ($\uparrow$) measures the fraction of correctly decoded payload bits, reported under: \emph{clean} (no augmentation) and \emph{noise average}. For images, the noise-average is mean over 17 medium-severity kornia augmentations~\cite{bui2025trustmark}: rotation, brightness, contrast, colour jitter, grayscale, Gaussian/motion/median/box blur, Gaussian noise, hue, posterize, RGB shift, saturation, sharpness, and JPEG. For videos, the noise-average also included video compression using codecs H.264, H.265/HEVC, VP9, and AV1 with quality ranging between CRF 18-36. 

\paragraph{Coexistence evaluation protocol.}
In all coexistence experiments, the first watermark (hereafter, WM1) which is applied to a clean cover image, then second watermark (hereafter, WM2) is embedded on top, and WM1 bit accuracy is measured from the doubly-watermarked image. The signpost can be applied as WM1 or WM2 depending upon if the other watermark is deployed with or without the access of signpost watermark. % This mirrors the deployment scenario where content may already carry an independently applied watermark when the signpost is overlaid.

% \begin{figure*}[t!]
%     \centering
%     \includegraphics[width=\textwidth]{figure/residuals.pdf}
%     \caption{Qualitative comparison of watermark residuals (amplified $\times 10$, gray = zero perturbation)
%     for four example images. \textbf{Ours} introduces sparse, minimal perturbations confined to
%     a small fraction of pixels, while PixelSeal distributes a smooth global texture, InvisMark and
%     TrustMark-P embed subtle but spatially dense patterns, and MaskWM applies a visible structured mask.
%     This sparsity explains our significantly higher solo PSNR of 52.4\,dB versus 37.9--47.4\,dB for baselines.}
%     \label{fig:residuals}
% \end{figure*}

\begin{figure*}[t!]
    \centering
    \begin{subfigure}{\textwidth}
        \centering
        \includegraphics[width=0.95\textwidth]{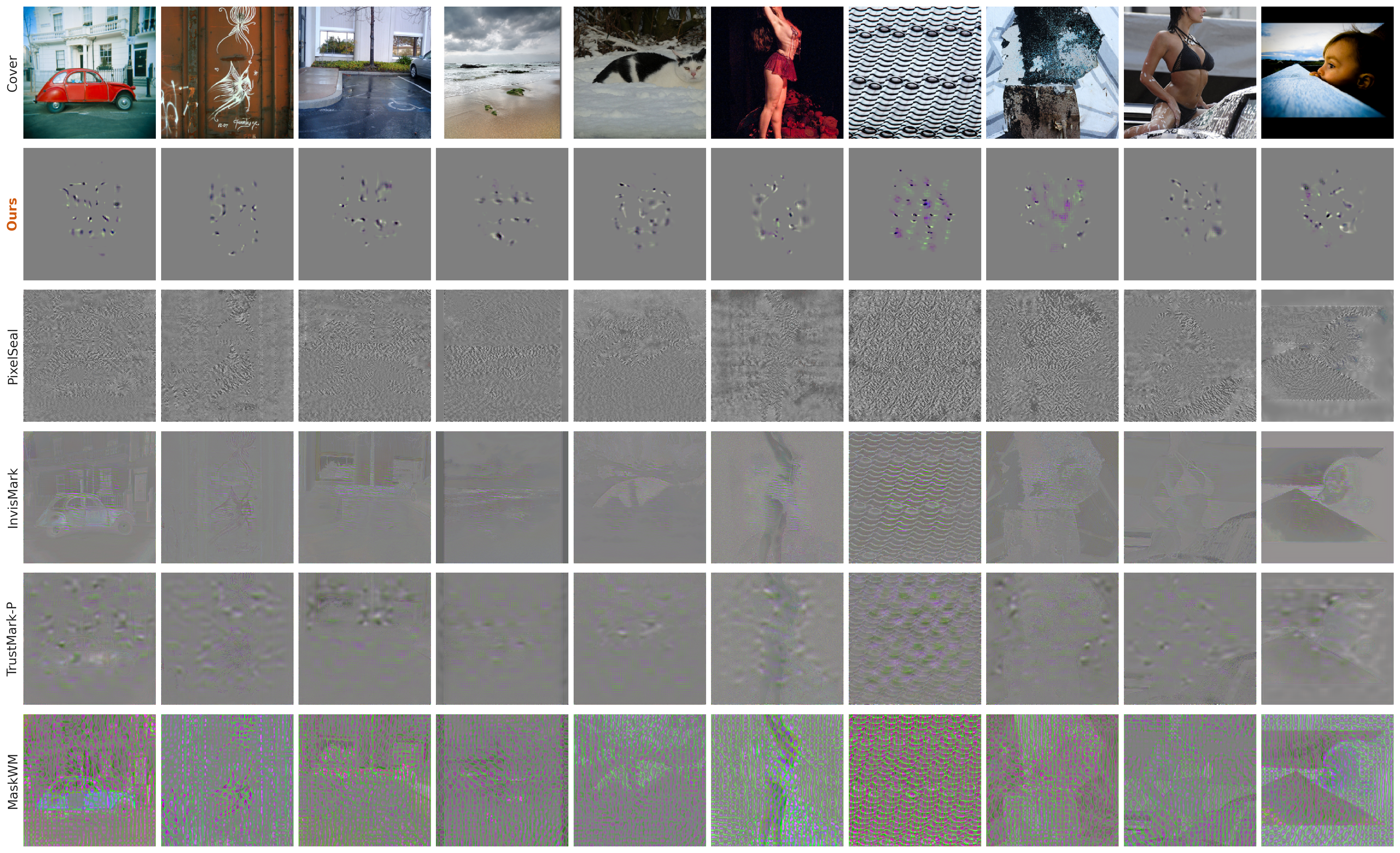}
        \caption{Image watermark residuals.}
        \label{fig:residuals_image}
    \end{subfigure}
    \\[1em]
    \begin{subfigure}{\textwidth}
        \centering
        \includegraphics[width=0.95\textwidth]{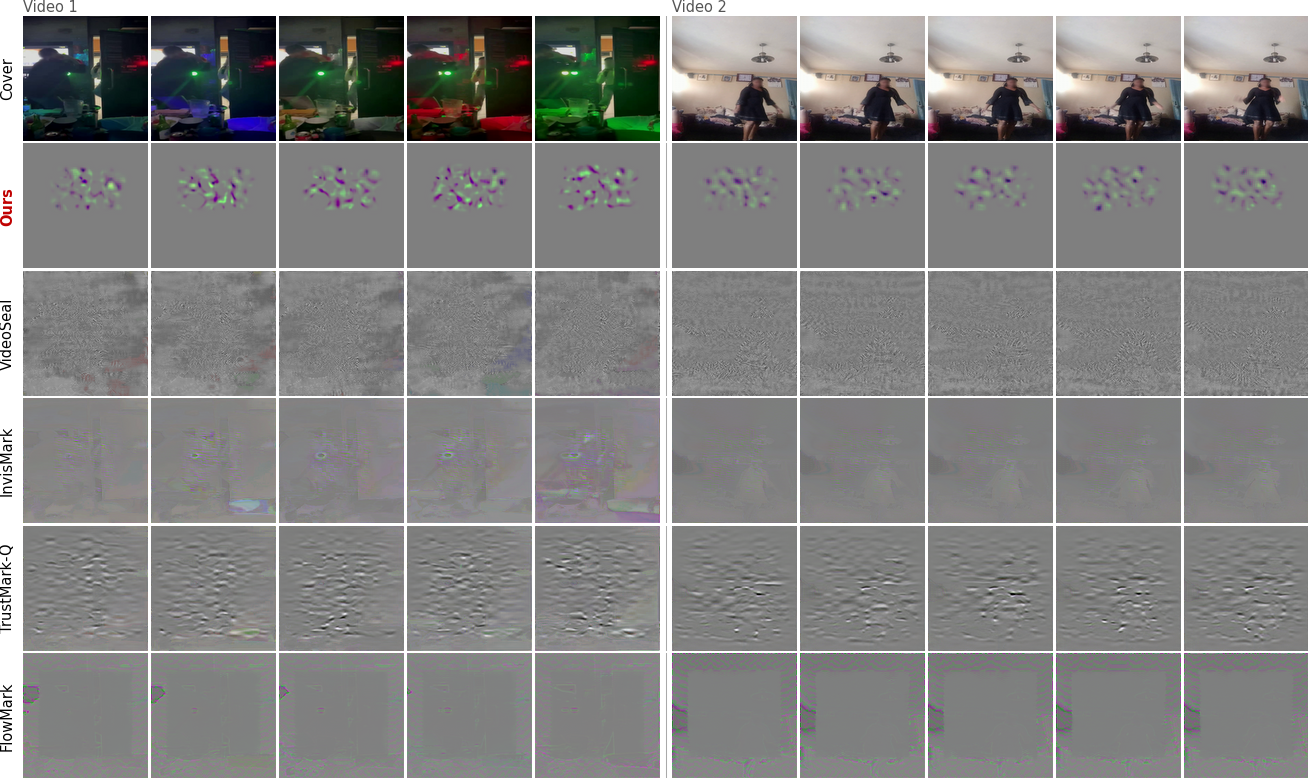}
        \caption{Video watermark residuals.}
        \label{fig:residuals_video}
    \end{subfigure}
    \caption{Qualitative comparison of watermark residuals (amplified $\times 10$, gray = zero perturbation). Shown here are residuals for (a) 10 image example and (b) 5 frames for two videos. In both cases, our signpost introduces sparse, minimal perturbations confined to 
    a small fraction of pixels, while other watermarking techniques distributes a smooth global texture. This sparsity explains our significantly higher solo PSNR of 52.4\,dB versus less than 50\,dB for others.}
    \label{fig:residuals}
\end{figure*}

% =====================================================================
%  4.2  QUALITY  (three tables merged into one)
% =====================================================================
\subsection{Watermark Quality}
\label{sec:quality}

Table~\ref{tab:quality} consolidates the watermark quality for image PSNR (top) and video PSNR/VMAF (bottom). Diagonal entries give the solo score; off-diagonal entries give the
score when WM1 (row) is followed by WM2 (column), measured against the unwatermarked original.

\begin{table}[t]
\centering
\caption{Watermark quality. \textbf{(a)}~Image PSNR (dB\,$\uparrow$).
\textbf{(b)}~Video quality reported as \emph{PSNR\,/\,VMAF} ($\uparrow$). Diagonal:
single watermark applied alone. Off-diagonal: WM1 (row) applied followed by WM2
(column). Best solo score per panel in \textbf{bold}.}
\label{tab:quality}
\scriptsize
\setlength{\tabcolsep}{4.5pt}

\begin{subtable}{\linewidth}
\centering
\caption{Image quality under coexistence -- PSNR (dB)}
\begin{tabular}{lcccccc}
\toprule
WM1\,$\downarrow$ / WM2\,$\rightarrow$ & PS & IM & MW & TM-P & ZOETROPE & Ours \\
\midrule
PS~\cite{souvcek2025pixel}       & 43.3 & 41.6 & 36.8 & 41.9 & 41.4 & 42.8 \\
IM~\cite{xu2025invismark}       & 41.3 & 46.3 & 37.0 & 43.0 & 42.5 & 45.2 \\
MW~\cite{hu2026mask}          & 36.8 & 37.3 & 37.9 & 37.3 & 37.0 & 37.7 \\
TM-P~\cite{bui2025trustmark}     & 41.8 & 43.0 & 37.2 & 47.4 & 43.0 & 46.1 \\
\hline
ZOETROPE~\cite{Collomosse-IEEECGA-2026}  & 41.4 & 42.5 & 37.0 & 43.2 & 45.8 & 45.0 \\
\textbf{Ours}             & 42.8 & 45.3 & 37.7 & 46.2 & 44.9 & \textbf{52.4} \\
\bottomrule
\end{tabular}
\end{subtable}

\vspace{0.6em}

\begin{subtable}{\linewidth}
\centering
\caption{Video quality under coexistence -- PSNR (dB)\,/\,VMAF}
\setlength{\tabcolsep}{3pt}
\scriptsize
\begin{tabular}{lcccccc}
\toprule
WM1\,$\downarrow$ / WM2\,$\rightarrow$ & VS & IM & FM & TM-Q & FM-32 & Ours \\
\midrule
VS~\cite{fernandez2024video}      & 45.3\,/\,99.1 & 43.6\,/\,98.7 & 41.8\,/\,89.9 & 41.1\,/\,97.8 & 40.6\,/\,87.9 & 44.5\,/\,98.7 \\
IM~\cite{xu2025invismark}      & 43.5\,/\,98.7 & 49.3\,/\,99.2 & 43.1\,/\,98.9 & 41.9\,/\,97.8 & 42.1\,/\,98.1 & 47.5\,/\,98.8 \\
FM~\cite{asnani2026flowmark}       & 43.8\,/\,89.9 & 43.4\,/\,99.0 & 48.5\,/\,99.4 & 40.8\,/\,98.0 & 43.0\,/\,94.2 & 44.2\,/\,99.0 \\
TM-Q~\cite{bui2025trustmark}    & 41.1\,/\,97.8 & 41.9\,/\,97.8 & 40.8\,/\,98.0 & 43.1\,/\,98.3 & 42.4\,/\,92.4 & 42.6\,/\,97.9 \\
\hline
FM-32~\cite{asnani2026flowmark}    & 38.7\,/\,90.0 & 41.9\,/\,98.1 & 43.2\,/\,95.1 & 41.6\,/\,92.4 & 48.1\,/\,99.1 & 44.6\,/\,98.4 \\
\textbf{Ours}            & 44.5\,/\,98.7 & 47.5\,/\,98.8 & 44.2\,/\,99.0 & 42.6\,/\,97.9 & 44.3\,/\,98.9 & \textbf{52.4\,/\,99.2} \\
\bottomrule
\end{tabular}
\end{subtable}
\vspace{-0.2cm}
\end{table}

\paragraph{Images.}
Our signpost attains the highest solo PSNR of $52.4$~dB, well above every image
baseline ($37.9$--$47.4$~dB). Figure~\ref{fig:residuals_image} (top) explains the gap: our
residual is sparse and spatially localized, whereas PixelSeal spreads a smooth global
texture, InvisMark and TrustMark-P embed dense low-amplitude patterns, and MaskWM
applies a structured, visible mask. When a second watermark is overlaid on ours, the
joint PSNR remains high, up to $46.2$~dB, with at most $\approx\!1.2$~dB below the corresponding solo values, versus up to $4.2$~dB when ZOETROPE is the base layer(with TrustMark-P). Our residual
therefore best preserves the quality of a subsequent embedding.

\paragraph{Video.}
The same trends hold in the temporal domain. Our video signpost again attains the
highest solo PSNR ($52.4$~dB), well above every video baseline ($43.1$--$49.3$~dB).
VMAF is far less discriminative here: all video watermarks are near-transparent
($98.3$--$99.4$ solo, ours at $99.2$), so PSNR is the more informative quality metric
in this regime. Qualitatively (Fig.~\ref{fig:residuals_video}, bottom), the video
residual is again sparse and spatially concentrated rather than a dense global
texture, closely resembling the image residuals in Fig.~\ref{fig:residuals_image}. We
attribute this cross-modal consistency to the shared signpost architecture, which
appears to induce similar residual patterns regardless of whether it is trained on
images or video.

The one pairing that visibly costs quality is FlowMark. Being mask-guided, FlowMark is
the video counterpart of MaskWM in the image experiments, and its residual dominates
the joint embedding: coexisting with FlowMark lowers the joint PSNR to $44.2$~dB, a
$4.3$~dB reduction from FlowMark's $48.5$~dB solo, and even the lighter FlowMark-32
variant drops to $44.3$~dB, $3.8$~dB below its own $48.1$~dB solo. Every other pairing
costs at most $1.8$~dB. VMAF, by contrast, remains high across all combinations
($97.9$--$99.0$), the largest reduction ($-1.3$) occurring when TrustMark-Q is applied
on top of our signpost, a negligible perceptual cost given the coexistence gains
reported next.

% =====================================================================
%  4.3  COEXISTENCE  (merged 4.3 + duplicated 4.4 + video 4.5)
% =====================================================================
\subsection{Watermark Coexistence}
\label{sec:coexist}

Tables~\ref{tab:coexist-clean} and~\ref{tab:coexist-kornia} report WM1 bit accuracy
when WM2 is applied on top, under clean and noise-average conditions respectively.
Each cell shows accuracy with the change $\delta$ (in brackets) relative to WM1
applied alone.

\begin{table}[t]

\caption{Coexistence under \emph{clean} (no-augmentation) conditions: WM1 bit accuracy
($\uparrow$) with WM2 overlaid. Cells show acc\,($\delta$), $\delta$ relative to WM1
alone. Rows = WM1; columns = WM2. \textbf{(a)}~image (top), \textbf{(b)}~video (bottom).
$^{\ast}$FM-32 = FlowMark without the coexistence objective.}
\label{tab:coexist-clean}
\scriptsize
\setlength{\tabcolsep}{2.5pt}
\renewcommand{\arraystretch}{1.1}
\resizebox{0.9\textwidth}{!}{%
\begin{subtable}{\linewidth}
\centering
\begin{tabular}{lcccccc}
\toprule
Technique & PS & IM & MW & TM-P & ZOETROPE & Ours \\
\midrule
PS~\cite{souvcek2025pixel}      & ---            & 1.000\,\chg{$-$.000} & 0.998\,\chg{$-$.002} & 1.000\,\chg{$-$.000} & 0.999\,\chg{$-$.001} & 1.000\,\chg{$-$.000} \\
IM~\cite{xu2025invismark}      & 0.954\,\chg{$-$.000} & ---            & 0.897\,\chg{$-$.056} & 0.939\,\chg{$-$.015} & 0.917\,\chg{$-$.037} & 0.952\,\chg{$-$.002} \\
MW~\cite{hu2026mask}         & 1.000\,\chg{$+$.000} & 1.000\,\chg{$+$.000} & ---            & 1.000\,\chg{$+$.000} & 1.000\,\chg{$+$.000} & 1.000\,\chg{$+$.000} \\
TM-P~\cite{bui2025trustmark}    & 0.984\,\chg{$-$.012} & 0.994\,\chg{$-$.002} & 0.932\,\chg{$-$.064} & ---            & 0.723\,\chg{$-$.273} & 0.988\,\chg{$-$.008} \\
\hline
ZOETROPE~\cite{Collomosse-IEEECGA-2026} & 0.999\,\chg{$-$.001} & 0.999\,\chg{$-$.000} & 0.995\,\chg{$-$.004} & 0.914\,\chg{$-$.085} & ---            & 0.998\,\chg{$-$.001} \\
\textbf{Ours}            & 0.992\,\chg{$-$.006} & 0.997\,\chg{$-$.001} & 0.993\,\chg{$-$.004} & 0.990\,\chg{$-$.008} & 0.871\,\chg{$-$.126} & ---            \\
\bottomrule
\end{tabular}
\end{subtable}
}
\vspace{0.6em}
\resizebox{0.9\textwidth}{!}{%
\begin{subtable}{\linewidth}
\centering
\begin{tabular}{lcccccc}
\toprule
Technique & VS & IM & FM & TM-Q & FM-32$^{\ast}$ & Ours \\
\midrule
VS~\cite{fernandez2024video}      & ---            & 1.000\,\chg{$-$.000} & 1.000\,\chg{$-$.000} & 1.000\,\chg{$-$.000} & 0.998\,\chg{$-$.002} & 1.000\,\chg{$-$.000} \\
IM~\cite{xu2025invismark}      & 0.955\,\chg{$-$.000} & ---            & 0.954\,\chg{$-$.001} & 0.930\,\chg{$-$.025} & 0.951\,\chg{$-$.004} & 0.953\,\chg{$-$.002} \\
FM~\cite{asnani2026flowmark}       & 1.000\,\chg{$+$.000} & 1.000\,\chg{$+$.000} & ---            & 1.000\,\chg{$+$.000} & 0.999\,\chg{$-$.001} & 1.000\,\chg{$+$.000} \\
TM-Q~\cite{bui2025trustmark}    & 0.979\,\chg{$-$.012} & 0.989\,\chg{$-$.002} & 0.966\,\chg{$-$.025} & ---            & 0.986\,\chg{$-$.005} & 0.991\,\chg{$-$.003} \\
\hline
FM-32~\cite{asnani2026flowmark}$^{\ast}$ & 0.998\,\chg{$-$.001} & 0.997\,\chg{$-$.002} & 0.998\,\chg{$-$.001} & 0.994\,\chg{$-$.005} & ---            & 0.997\,\chg{$-$.002} \\
\textbf{Ours}            & 0.997\,\chg{$-$.001} & 0.997\,\chg{$-$.001} & 0.997\,\chg{$+$.000} & 0.991\,\chg{$-$.003} & 0.994\,\chg{$-$.003} & ---            \\
\bottomrule
\end{tabular}
\end{subtable}
}
\vspace{-0.2cm}
\end{table}

\begin{table}[t]
\caption{Watermark coexistence: bit accuracy of first watermark (WM1) when a second watermark (WM2) is applied on top, averaged over the noise setting described in Section~\ref{sec:setup}. Each cell shows \texttt{acc (}\(\pm\delta\)\texttt{)} where \(\delta\) is the change relative to WM1 applied alone (diagonal). Rows = WM1 applied first; columns = WM2 applied on top. Layout as in
Table~\ref{tab:coexist-clean}.}
\label{tab:coexist-kornia}
\scriptsize
\setlength{\tabcolsep}{2.5pt}
\renewcommand{\arraystretch}{1.1}
\resizebox{0.9\textwidth}{!}{%
\begin{subtable}{\linewidth}
\begin{tabular}{lcccccc}
\toprule
Technique & PS & IM & MW & TM-P & ZOETROPE & Ours \\
\midrule
PS~\cite{souvcek2025pixel}      & ---            & 0.976\,\chg{$-$.000} & 0.971\,\chg{$-$.005} & 0.973\,\chg{$-$.002} & 0.971\,\chg{$-$.004} & 0.973\,\chg{$-$.002} \\
IM~\cite{xu2025invismark}      & 0.866\,\chg{$-$.003} & ---            & 0.771\,\chg{$-$.098} & 0.852\,\chg{$-$.017} & 0.837\,\chg{$-$.032} & 0.865\,\chg{$-$.004} \\
MW~\cite{hu2026mask}         & 0.982\,\chg{$-$.002} & 0.983\,\chg{$-$.001} & ---            & 0.982\,\chg{$-$.003} & 0.983\,\chg{$-$.002} & 0.984\,\chg{$-$.001} \\
TM-P~\cite{bui2025trustmark}    & 0.920\,\chg{$-$.021} & 0.934\,\chg{$-$.007} & 0.867\,\chg{$-$.075} & ---            & 0.687\,\chg{$-$.254} & 0.929\,\chg{$-$.013} \\
\hline
ZOETROPE~\cite{Collomosse-IEEECGA-2026} & 0.974\,\chg{$-$.006} & 0.978\,\chg{$-$.001} & 0.963\,\chg{$-$.016} & 0.879\,\chg{$-$.100} & ---            & 0.972\,\chg{$-$.007} \\
\textbf{Ours}            & 0.967\,\chg{$-$.011} & 0.976\,\chg{$-$.002} & 0.969\,\chg{$-$.009} & 0.962\,\chg{$-$.016} & 0.831\,\chg{$-$.147} & ---            \\
\bottomrule
\end{tabular}
\end{subtable}
}
\vspace{0.6em}
\renewcommand{\arraystretch}{1.1}
\resizebox{0.9\textwidth}{!}{%
\begin{subtable}{\linewidth}
\centering
\begin{tabular}{lcccccc}
\toprule
Technique & VS & IM & FM & TM-Q & FM-32 & Ours \\
\midrule
VS~\cite{fernandez2024video}      & ---            & 0.933\,\chg{$+$.000} & 0.932\,\chg{$-$.001} & 0.931\,\chg{$-$.001} & 0.930\,\chg{$-$.003} & 0.933\,\chg{$-$.000} \\
IM~\cite{xu2025invismark}      & 0.829\,\chg{$-$.001} & ---            & 0.822\,\chg{$-$.007} & 0.812\,\chg{$-$.017} & 0.818\,\chg{$-$.012} & 0.828\,\chg{$-$.002} \\
FM~\cite{asnani2026flowmark}       & 0.911\,\chg{$-$.000} & 0.910\,\chg{$-$.001} & ---            & 0.911\,\chg{$+$.001} & 0.916\,\chg{$+$.005} & 0.911\,\chg{$-$.000} \\
TM-Q~\cite{bui2025trustmark}    & 0.870\,\chg{$-$.013} & 0.880\,\chg{$-$.003} & 0.861\,\chg{$-$.023} & ---            & 0.865\,\chg{$-$.018} & 0.945\,\chg{$-$.003} \\
\hline
FM-32~\cite{asnani2026flowmark} & 0.917\,\chg{$-$.001} & 0.910\,\chg{$-$.008} & 0.916\,\chg{$-$.002} & 0.905\,\chg{$-$.013} & ---            & 0.915\,\chg{$-$.003} \\
\textbf{Ours}            & 0.927\,\chg{$-$.004} & 0.928\,\chg{$-$.002} & 0.930\,\chg{$-$.001} & 0.965\,\chg{$-$.001} & 0.918\,\chg{$-$.012} & ---            \\
\bottomrule
\end{tabular}
\end{subtable}
}
\end{table}

\paragraph{Clean conditions.}
Most pairs already coexist well without explicit training. PixelSeal and MaskWM are
especially robust as WM1, holding bit accuracy $\ge 0.99$ regardless of the overlaid
signpost. TrustMark-P is the most sensitive, dropping to $0.723$ under ZOETROPE but
recovering to $0.988$ under ours; InvisMark similarly improves from $0.917$ (ZOETROPE)
to $0.952$ (ours). The $\delta$ values are more informative than the raw numbers:
with the signpost as WM2, InvisMark and our method look similar, but that similarity
holds only in this direction---as WM1, InvisMark's solo accuracy ($0.954$) is well
below ours ($0.998$) and we achieve
it at markedly higher quality ($52.4$ vs.\ $46.3$~dB).

Video coexistence is even more forgiving. Under clean conditions the video panel is
essentially saturated: every WM1 retains bit accuracy within $0.025$ of its solo value
under any overlay, and no video watermark struggles to coexist with the video signpost.
VS and FM both exceed $0.998$ as WM1 under any overlay. The most telling
case is FM-32, a signpost built on the FlowMark architecture but deliberately
held out of our coexistence training: it coexists with our video signpost with  accuracy holding at $0.994$ ($-0.003$) under an FM-32 overlay while FM-32
itself stays at $0.997$ ($-0.002$) under ours. We hypothesize that as FM-32 shares the mask-guided FlowMark residual it already sits in complementary regions to our signpost. 
This contrasts with the image setting, where the analogous held-out
signpost, ZOETROPE, is the hardest case: TrustMark-P collapses to $0.723$
under a ZOETROPE overlay, and even our own signpost drops to $0.871$. These failures are not a concern, as two competing
signposts sharing a single image is not a design requirement and two signposts are never deployed together.

\paragraph{Under augmentation.}
Augmentation sharpens the differences. In images, the most pronounced gain is for
TrustMark-P: ZOETROPE leaves it at $0.687$ noise-average accuracy, while our
decoder-aware model raises it to $0.929$ ($+0.24$); InvisMark improves from $0.837$ to
$0.865$. PixelSeal and MaskWM stay near the ceiling ($\ge 0.97$) for every method,
reflecting their intrinsic robustness. The same pattern transfers to video, where
TrustMark-Q is again the hardest case: it retains $0.945$ bit accuracy under our
signpost overlay, far above the $0.861$--$0.880$ it retains under every other video
overlay, and in the reverse direction our own signpost reaches $0.965$ under
TrustMark-Q, its strongest coexistence result against any secondary video watermark
(vs.\ $0.927$ and $0.930$ under VideoSeal and FlowMark overlays). Decoder-aware
training is thus most beneficial precisely for the watermark family that struggles most
under naive overlay.

% =====================================================================
%  4.4  ABLATION
% =====================================================================
\subsection{Ablation Study}
\label{sec:ablation}

Table~\ref{tab:ablation} reports the signpost-first (SP-first) scenario: the signpost
is applied first to the clean cover and a secondary watermark is overlaid, measuring
noise-average SP bit accuracy. We ablate each frozen decoder across three groups:
\texttt{noCoexist} (coexistence loss disabled), four single-decoder variants, and four
leave-one-out variants, on per-image noise-average accuracy ($N{=}1000$).

\begin{table*}[t!]
\caption{Ablation (SP-first) for image/video (top/bottom): signpost applied first followed by WM2 applied on top. Columns show noise-average SP bit accuracy when each secondary WM is overlaid. \textit{Solo}: SP alone (no overlay). \ding{51}/\ding{55}: coexistence trained for that frozen decoder. Right: solo PSNR\,(dB).}
\label{tab:ablation}
\scriptsize
\setlength{\tabcolsep}{4pt}
\renewcommand{\arraystretch}{1.2}
\resizebox{0.95\textwidth}{!}{%
\begin{tabular}{lcccc|c|cccc|c}
\toprule
& \multicolumn{4}{c|}{\textbf{Coexist training}} & \textbf{Solo SP} & \multicolumn{4}{c|}{\textbf{SP bit acc. with WM2} ($\uparrow$)} & \textbf{PSNR} \\
\cmidrule(lr){2-5}\cmidrule(lr){6-6}\cmidrule(lr){7-10}\cmidrule(lr){11-11}
\textbf{Variant} & \textbf{PS} & \textbf{IM} & \textbf{TM-P} & \textbf{MW} & \textbf{acc.} & \textbf{PS} & \textbf{IM} & \textbf{TM-P} & \textbf{MW} & \textbf{(dB)} $\uparrow$ \\
\midrule
noCoexist & \textcolor{red!60!black}{\ding{55}} & \textcolor{red!60!black}{\ding{55}} & \textcolor{red!60!black}{\ding{55}} & \textcolor{red!60!black}{\ding{55}} & 0.957 & 0.949 & 0.954 & 0.841 & 0.911 & 53.0 \\
\midrule
Only PS & \textcolor{teal}{\ding{51}} & \textcolor{red!60!black}{\ding{55}} & \textcolor{red!60!black}{\ding{55}} & \textcolor{red!60!black}{\ding{55}} & 0.958 & 0.957 & 0.956 & 0.849 & 0.927 & 52.7 \\
Only IM & \textcolor{red!60!black}{\ding{55}} & \textcolor{teal}{\ding{51}} & \textcolor{red!60!black}{\ding{55}} & \textcolor{red!60!black}{\ding{55}} & 0.959 & 0.956 & 0.958 & 0.868 & 0.919 & 52.3 \\
Only TM-P & \textcolor{red!60!black}{\ding{55}} & \textcolor{red!60!black}{\ding{55}} & \textcolor{teal}{\ding{51}} & \textcolor{red!60!black}{\ding{55}} & 0.951 & 0.947 & 0.947 & 0.944 & 0.890 & 52.6 \\
Only MW & \textcolor{red!60!black}{\ding{55}} & \textcolor{red!60!black}{\ding{55}} & \textcolor{red!60!black}{\ding{55}} & \textcolor{teal}{\ding{51}} & 0.947 & 0.937 & 0.944 & 0.874 & 0.944 & 49.9 \\
\midrule
w/o PS & \textcolor{red!60!black}{\ding{55}} & \textcolor{teal}{\ding{51}} & \textcolor{teal}{\ding{51}} & \textcolor{teal}{\ding{51}} & 0.944 & 0.939 & 0.942 & 0.886 & 0.909 & \textbf{54.0} \\
w/o IM & \textcolor{teal}{\ding{51}} & \textcolor{red!60!black}{\ding{55}} & \textcolor{teal}{\ding{51}} & \textcolor{teal}{\ding{51}} & 0.956 & 0.953 & 0.954 & 0.909 & 0.931 & 53.4 \\
w/o TM-P & \textcolor{teal}{\ding{51}} & \textcolor{teal}{\ding{51}} & \textcolor{red!60!black}{\ding{55}} & \textcolor{teal}{\ding{51}} & 0.950 & 0.945 & 0.948 & 0.822 & 0.923 & 53.2 \\
w/o MW & \textcolor{teal}{\ding{51}} & \textcolor{teal}{\ding{51}} & \textcolor{teal}{\ding{51}} & \textcolor{red!60!black}{\ding{55}} & 0.959 & 0.956 & 0.957 & 0.928 & 0.925 & 52.8 \\
\midrule
\textbf{Ours} & \textcolor{teal}{\ding{51}} & \textcolor{teal}{\ding{51}} & \textcolor{teal}{\ding{51}} & \textcolor{teal}{\ding{51}} & \textbf{0.977} & \textbf{0.965} & \textbf{0.974} & \textbf{0.960} & \textbf{0.967} & 52.3 \\
\midrule
\midrule
\textbf{Variant} & \textbf{VS} & \textbf{IM} & \textbf{TM-Q} & \textbf{FM} & \textbf{acc.} & \textbf{VS} & \textbf{IM} & \textbf{TM-Q} & \textbf{FM} & \textbf{(dB)} $\uparrow$ \\
\midrule
noCoexist & \textcolor{red!60!black}{\ding{55}} & \textcolor{red!60!black}{\ding{55}} & \textcolor{red!60!black}{\ding{55}} & \textcolor{red!60!black}{\ding{55}} & 0.921 & 0.918 & 0.919 & 0.914 & 0.920 & 52.0 \\
\midrule
Only VS & \textcolor{teal}{\ding{51}} & \textcolor{red!60!black}{\ding{55}} & \textcolor{red!60!black}{\ding{55}} & \textcolor{red!60!black}{\ding{55}} & 0.920 & 0.919 & 0.920 & 0.912 & 0.921 & 51.8 \\
Only IM & \textcolor{red!60!black}{\ding{55}} & \textcolor{teal}{\ding{51}} & \textcolor{red!60!black}{\ding{55}} & \textcolor{red!60!black}{\ding{55}} & 0.924 & 0.921 & 0.923 & 0.916 & 0.924 & 51.3 \\
Only TM-Q & \textcolor{red!60!black}{\ding{55}} & \textcolor{red!60!black}{\ding{55}} & \textcolor{teal}{\ding{51}} & \textcolor{red!60!black}{\ding{55}} & 0.926 & 0.923 & 0.925 & 0.924 & 0.925 & 51.5 \\
Only FM & \textcolor{red!60!black}{\ding{55}} & \textcolor{red!60!black}{\ding{55}} & \textcolor{red!60!black}{\ding{55}} & \textcolor{teal}{\ding{51}} & 0.911 & 0.907 & 0.907 & 0.900 & 0.919 & 51.6 \\
\midrule
w/o VS & \textcolor{red!60!black}{\ding{55}} & \textcolor{teal}{\ding{51}} & \textcolor{teal}{\ding{51}} & \textcolor{teal}{\ding{51}} & 0.911 & 0.908 & 0.910 & 0.908 & 0.912 & \textbf{52.5} \\
w/o IM & \textcolor{teal}{\ding{51}} & \textcolor{red!60!black}{\ding{55}} & \textcolor{teal}{\ding{51}} & \textcolor{teal}{\ding{51}} & 0.919 & 0.915 & 0.916 & 0.915 & 0.918 & 52.4 \\
w/o TM-Q & \textcolor{teal}{\ding{51}} & \textcolor{teal}{\ding{51}} & \textcolor{red!60!black}{\ding{55}} & \textcolor{teal}{\ding{51}} & 0.917 & 0.914 & 0.916 & 0.912 & 0.918 & 52.3 \\
w/o FM & \textcolor{teal}{\ding{51}} & \textcolor{teal}{\ding{51}} & \textcolor{teal}{\ding{51}} & \textcolor{red!60!black}{\ding{55}} & 0.920 & 0.919 & 0.918 & 0.918 & 0.920 & 52.4 \\
\midrule
\textbf{Ours} & \textcolor{teal}{\ding{51}} & \textcolor{teal}{\ding{51}} & \textcolor{teal}{\ding{51}} & \textcolor{teal}{\ding{51}} & \textbf{0.971} & \textbf{0.928} & \textbf{0.930} & \textbf{0.965} & \textbf{0.930} & 52.4 \\
\bottomrule
\end{tabular}%
}
\vspace{-0.2cm}
\end{table*}

\paragraph{Overall.}
Across both modalities the full model dominates every ablated variant. For images,
\texttt{noCoexist} degrades most under TrustMark-P overlay, falling from a solo $0.957$
to $0.841$ ($-0.116$), with MaskWM the second-hardest overlay; our full model raises SP
accuracy under every secondary watermark and lifts solo accuracy from $0.957$ to $0.977$. The video ablation
follows the same shape: TrustMark-Q is again the hardest overlay ($0.914$ for
\texttt{noCoexist}) and benefits most from decoder-aware training, rising to $0.965$ in
the full model ($+0.051$), while solo accuracy improves from $0.921$ to $0.971$. In both
settings coexistence training simultaneously sharpens the signpost's own intrinsic
robustness, and the TrustMark family---TrustMark-P for images, TrustMark-Q for
video---is consistently where explicit supervision pays off most.

\paragraph{Single-decoder specialization.}
Single-decoder variants reveal antagonistic specialization, most visibly in the image
setting. \emph{Only TM} raises TrustMark-P-overlay accuracy from $0.841$ to $0.944$
($+0.103$) but reduces MaskWM-overlay robustness to $0.890$; conversely \emph{Only MW}
improves MaskWM overlay while degrading PixelSeal- and InvisMark-overlay robustness.
Each frozen decoder shapes a residual tuned to one interference type at the expense of
others. In video the single-decoder effects are muted: no individual decoder reproduces
the full model's accuracy of $0.965$. This indicates that the
video decoders act synergistically, and the strongest coexistence behavior emerges only
when all of them are trained together.

\paragraph{Which decoders matter.}
The leave-one-out variants confirm that every decoder contributes: removing any single
one lowers overall overlay accuracy relative to the full model. For images, TrustMark-P
is the most important to retain, removing it (\emph{w/o TM}) fails to help
TrustMark-P coexistence ($0.822$), dropping the accuracy below the \texttt{noCoexist}
baseline of $0.841$, and produces the largest mean accuracy loss across the four
overlays ($0.057$, vs.\ $0.048$, $0.030$, $0.025$ for w/o~PS, w/o~IM, w/o~MW). Whereas,
MaskWM is the most dispensable (\emph{w/o MW} stays within $0.01$--$0.04$ of the full
model and retains the best solo accuracy, $0.959$). Video carries the same
qualitative message with a different ranking: VideoSeal is the most valuable decoder to
keep (\emph{w/o VS} incurs the largest overall drop and the lowest solo accuracy,
$0.911$), and FlowMark the most dispensable (\emph{w/o FM} remains closest to the full
model at $0.920$ solo). 

%% file: sections/05_conclusion.tex
\section{Conclusion}
\label{sec:conclusion}

We presented a decoder-aware training framework for signpost watermarks that coexist with independently developed watermarking systems across both images and video. By placing frozen secondary decoders directly in the training loop, our method transforms coexistence from an emergent property of independently trained systems into an explicit optimization objective: the signpost encoder learns a residual that survives the overlay of each frozen watermarker, encouraging spatial and frequency complementarity. We further show that the coexistence behavior previously observed only for images extends to the temporal domain, providing the first evidence that independently developed video watermarks can coexist and that this coexistence can likewise be improved through explicit optimization. Our signpost achieves the highest perceptual quality among the evaluated methods, reaching a solo PSNR of $52.4$~dB for both images and video and the highest solo video VMAF ($99.2$). Decoder-aware training delivers its largest gains for the most interference-sensitive watermark family, TrustMark, improving augmentation-averaged coexistence by up to $+0.24$ bit accuracy over the previous signpost baseline while maintaining this high visual quality.

These results support the practical feasibility of a standardized `signpost' watermark as a lightweight routing signal into a registry of independently developed watermarking systems. Such a signpost would enable interoperable watermarking for content authenticity and IP rights signalling without requiring consensus on a single shared watermarking algorithm. Future work includes understanding robustness under adaptive attacks that explicitly target doubly-watermarked signals.  A practical next step could be for a suitable consortium or standards body to consider the merit of agreeing a lightweight payload format, for example a integer indexing a public list of provenance watermarking methods such as the C2PA `soft binding algorithm list' \cite{c2pa}. The proposed signpost would thus provide an interoperability layer, allowing independent image and video watermarking technologies to participate in a shared provenance and IP rights ecosystem.